%% file: main.tex
\documentclass[letterpaper, 10 pt, conference]{ieeeconf}  %

\IEEEoverridecommandlockouts                              %

\let\labelindent\relax

\usepackage[para]{footmisc}
\usepackage{wrapfig}
\usepackage{capt-of}
\usepackage[pagebackref=true,breaklinks=true,colorlinks=true,bookmarks=false,citecolor=blue]{hyperref}

\usepackage{graphics,enumitem} %
\usepackage{epsfig} %
\usepackage{mathptmx} %
\usepackage{tabularx}
\usepackage{times} %
\usepackage{amsmath} %
\usepackage{multicol}
\usepackage{amssymb}  %
\usepackage{booktabs}
\usepackage{multirow}
\usepackage{algpseudocode}
\usepackage{algorithm}
\usepackage{gensymb}
\usepackage{amssymb}
\usepackage{makecell}
\usepackage{caption}
\usepackage{subcaption}
\usepackage{url}
\usepackage{multicol}
\usepackage{wrapfig,lipsum,booktabs}
\usepackage{color}
\usepackage{cite}
\usepackage{balance}
\usepackage{physics}
\usepackage{float}
\usepackage{listings}
\usepackage{xspace}
\usepackage{dsfont}
\usepackage{longtable}
\usepackage{txfonts}
\usepackage[pagebackref=true,breaklinks=true,colorlinks=true,bookmarks=false,citecolor=blue]{hyperref}
\usepackage[dvipsnames]{xcolor}
\hypersetup{
colorlinks=true,
linkcolor=blue,
filecolor=magenta,      
urlcolor=blue,
citecolor=blue
}
\newcommand{\ours}{ManipGen\xspace}
\definecolor{Gray}{gray}{0.9}

\title{\LARGE \bf
Local Policies Enable Zero-shot Long-horizon Manipulation
}

\author{Murtaza Dalal$^{\ast}$$^{1}$\qquad Min Liu$^{\ast}$$^{1}$\qquad Walter Talbott$^{2}$ \qquad Chen Chen$^{2}$ \\
Deepak Pathak$^{1}$ \qquad Jian Zhang$^{2}$ \qquad Ruslan Salakhutdinov$^{1}$\\
\small $^{1}$Carnegie Mellon University, $^{2}$Apple
}

\begin{document}
\makeatletter
\let\@oldmaketitle\@maketitle%
\renewcommand{\@maketitle}{\@oldmaketitle%
\includegraphics[width=1\linewidth]{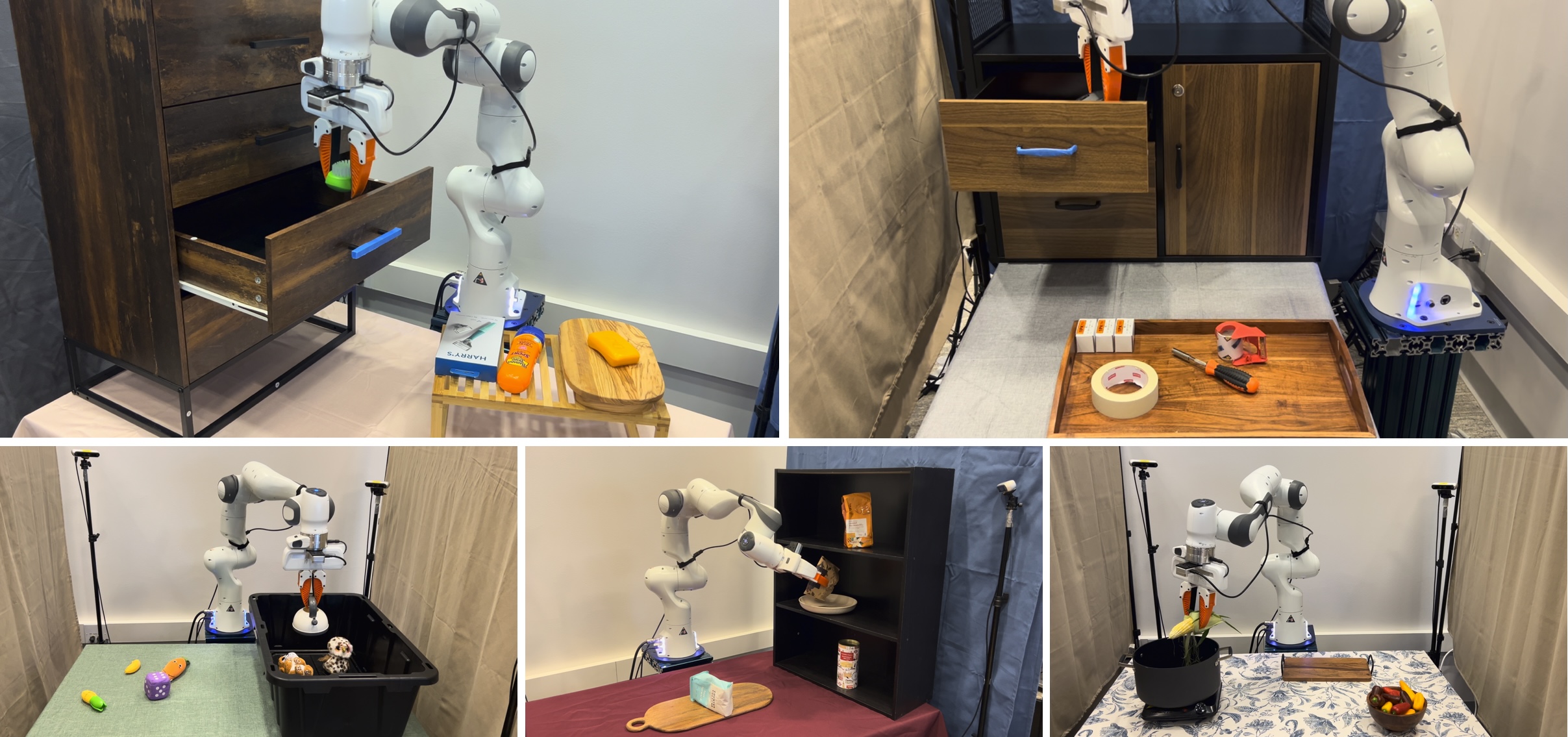}
  \centering
  \vspace{-12pt}
  \captionof{figure}{\small \textbf{Zero-shot Long-horizon Manipulation} Our approach trains a library of generalist manipulation skills in simulation and transfers them zero-shot to long-horizon manipulation tasks. We show a single, text-conditioned agent can manipulate unseen objects, in arbitrary poses and scene configurations, across long-horizons in the real world, solving challenging manipulation tasks with complex obstacles. }
    \vspace{-7pt}
  \label{fig:teaser}
  }
\makeatother
\maketitle
\thispagestyle{empty}
\pagestyle{empty}

\begin{abstract}
Sim2real for robotic manipulation is difficult due to the challenges of simulating complex contacts and generating realistic task distributions. To tackle the latter problem, we introduce \ours, which leverages a new class of policies for sim2real transfer: local policies. 
Locality enables a variety of appealing properties including invariances to absolute robot and object pose, skill ordering, and global scene configuration.
We combine these policies with foundation models for vision, language and motion planning and demonstrate SOTA zero-shot performance of our method to Robosuite benchmark tasks in simulation (97\%). We transfer our local policies from simulation to reality and observe they can solve unseen long-horizon manipulation tasks with up to 8 stages with significant pose, object and scene configuration variation. \ours outperforms SOTA approaches such as SayCan, OpenVLA, LLMTrajGen and VoxPoser across 50 real-world manipulation tasks by 36\%, 76\%, 62\% and 60\% respectively. 
Video results at \texttt{\href{https://mihdalal.github.io/manipgen/}{mihdalal.github.io/manipgen}}
\end{abstract}

\let\thefootnote\relax\footnote{\textsuperscript{*}equal contribution.}
\vspace{-10pt}
\section{Introduction}
\label{sec:intro}

How can we develop generalist robot systems that plan, reason, and interact with the world like humans? Tasks that humans solve during their daily lives, such as those shown in Figure~\ref{fig:teaser}, are incredibly challenging for existing robotics approaches. Cleaning the table, organizing the shelf, putting items away inside drawers, etc. are complex, long-horizon problems that require the robot to act capably and consistently over an extended period of time. 
Furthermore, such a generalist robot should be able to do so without requiring task-specific engineering effort or demonstrations.
Although large-scale data-driven learning has produced generalists for vision and language~\cite{openai2023gpt4}, such models don't yet exist in robotics due to the challenges of scaling data collection. It often takes significant manual labor cost and years of effort to just collect datasets on the order of 100K-1M trajectories~\cite{collaboration2023open, khazatsky2024droid,ebert2021bridge,walke2023bridgedata}. Consequently, generalization is limited, often to within centimeters of an object's pose for complex tasks~\cite{zhao2023learning,fu2024mobile}.

Instead, we seek to use a large-scale approach via simulation-to-reality (sim2real) transfer, a cost-effective technique for generating vast datasets that has enabled training generalist policies for locomotion which can traverse complex, unstructured terrain~\cite{lee2020learning,kumar2021rma,agarwal2023legged,zhuang2023robot,cheng2023parkour,hoeller2024anymal}. While sim2real transfer has shown success in industrial manipulation tasks~\cite{tang2023industreal,tang2024automate,jiang2024transic}, including with high-dimensional hands~\cite{akkaya2019solving,handa2022dextreme,lum2024dextrah,chen2022visual}, these efforts often involve training and testing on the same task in simulation. Can we extend sim2real to open-world manipulation, where robots need to solve any task from text instruction?
\setcounter{figure}{1}
\begin{figure*}[t]
\includegraphics[width=\textwidth]{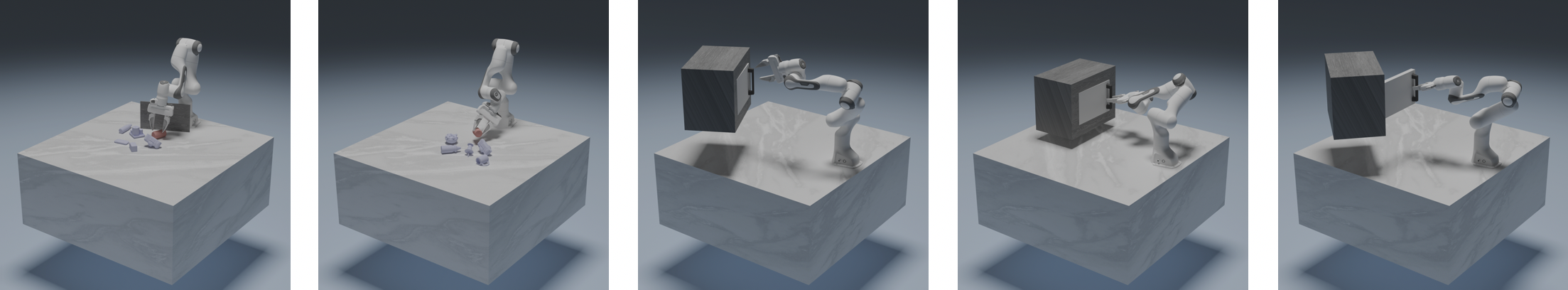}
\vspace{-14pt}
\caption{\small \textbf{Training Environments} We train local policies (left to right) on picking, placing, handle grasping, opening and closing.}
\vspace{-17pt}
\label{fig:sim vis}
\end{figure*}
The core bottlenecks are: 1) accurately simulating contact dynamics~\cite{todorov2012mujoco} -- for which strategies such as domain randomization~\cite{akkaya2019solving,andrychowicz2020learning}, SDF  contacts~\cite{narang2022factory,tang2023industreal,tang2024automate}, and real world corrections~\cite{jiang2024transic} have shown promise, 2) generating all possible scene and task configurations to ensure trained policies generalize and 3) acquiring long-horizon behaviors themselves, which may require potentially intractable amounts of data for as the horizon grows.

To address points 2) and 3), our solution is to note that for many manipulation tasks of interest, the skill can be simplified to two steps: achieving a pose near a target object, then performing manipulation. The key idea is that of \textit{locality of interaction}. 
Policies that observe and act in a region local to the target object of interest are by construction: 
\newline $\bullet$ \textbf{absolute pose invariant}: they reason about a much smaller set of relative poses between the objects and the robot.
\newline $\bullet$ \textbf{skill order invariant}: transition from the termination set of one policy and initiation set of the next via motion planning.
\newline $\bullet$ \textbf{scene configuration invariant}: they solely observe the local region around the point of interaction. 

We propose a novel approach that leverages the strong generalization capabilities of existing foundation models such as Visual Language Models (VLMs) for decomposing tasks into sub-problems~\cite{openai2023gpt4}, processing and understanding scenes~\cite{ren2024grounded} and planning collision-avoidant motions~\cite{dalal2024neuralmp}. 
Specifically, given a text prompt, our approach outputs a plan to solve the task (using a VLM), estimates where to go and moves the robot accordingly (using motion planning) and deploys local policies to perform interaction. As a result, a simple scene generation approach (Fig.~\ref{fig:sim vis}) can produce strong transfer results across many manipulation tasks (Fig.~\ref{fig:teaser}).

\looseness=-1
Our contribution is an approach to training agents at scale solely in simulation that are capable of solving a vast set of long-horizon manipulation tasks in the real world \textit{zero-shot}. Our method generalizes to unseen objects, poses, receptacles and skill order configurations. To do so, our method, \ours, 1) introduces a novel policy class for sim2real transfer 2) proposes techniques for training policies at scale in simulation 3) and deploys policies via integration with VLMs and motion planners. 
We perform a thorough, real world evaluation of \ours on \textbf{50} long-horizon manipulation tasks in \textbf{five} environments with up to \textbf{8} stages, achieving a success rate of \textbf{76\%}, outperforming SayCan, OpenVLA, LLMTrajGen and VoxPoser by \textbf{36\%}, \textbf{76\%}, 
 \textbf{62\%} and \textbf{60\%}.

\begin{figure*}[h!]
    \centering
    \vspace{5pt}
    \includegraphics[width=\linewidth]{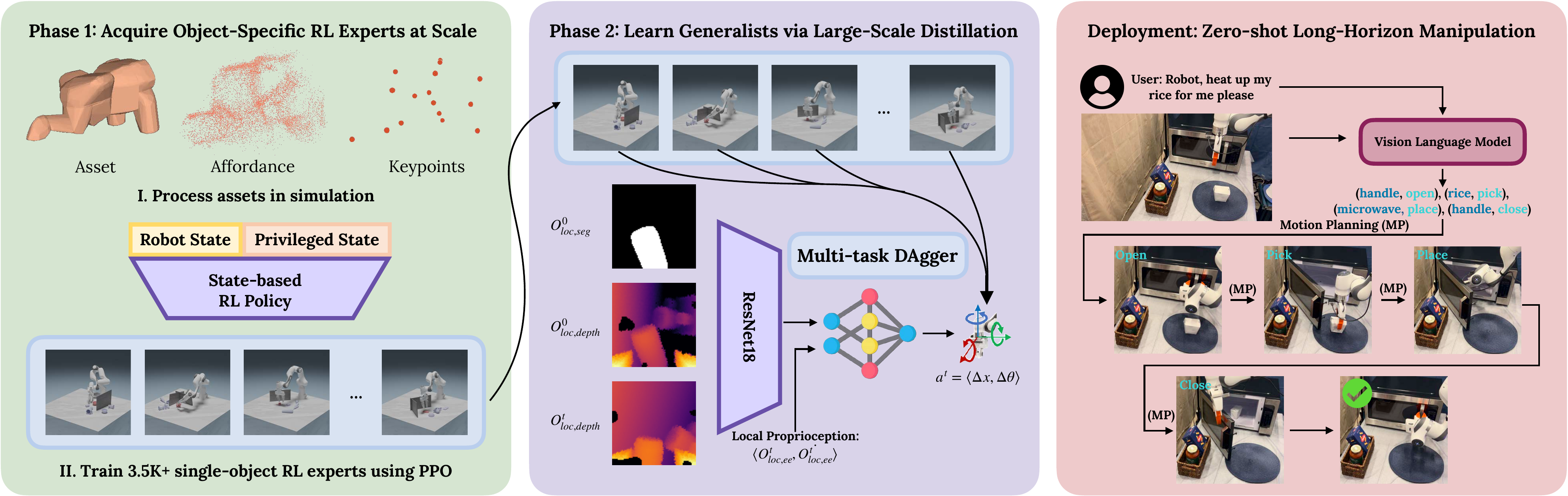}
    \vspace{-15pt}
    \caption{\small \textbf{\ours Method Overview} (\textit{left}) Train 1000s of RL experts in simulation using PPO (\textit{middle}) Distill single-task RL experts into generalist visuomotor policies via DAgger (\textit{right}) Text-conditioned long-horizon manipulation via task decomposition (VLM), pose estimation and goal reaching (Motion Planning) and sim2real transfer of local policies}
    \label{fig:main fig}
    \vspace{-18pt}
\end{figure*}

\vspace{-1pt}
\section{Related Work}
\label{sec:related work}
\looseness=-1\noindent \textbf{Long-horizon Robotic Manipulation}
Sense-Plan-Act (SPA) has been explored extensively over the past 50 years~\cite{center1984shakey,paul1981robot,whitney1972mathematics,vukobratovic1982dynamics,kappler2018real,murphy2019introduction}. Traditionally, SPA assumes access to accurate state estimation, a well-defined model of the environment and low-level control primitives. SPA, while capable of generalizing to a broad set of tasks, can require manual engineering and systems effort to set up~\cite{garrett2020online}, struggles with contact-rich interactions~\cite{mason2001mechanics,whitney2004mechanical} and fails due to state-estimation errors~\cite{kaelbling2013integrated}. By contrast, our method can be deployed to new tasks using generalist models which have minimal setup cost, train polices for contact-rich interactions and handle state-estimation issues by training with significant local randomization.

\noindent \textbf{Zero/Few-shot Manipulation Using Foundation Models}
The robotics community has begun to investigate VLM's capabilities for controlling robots in a zero/few-shot manner~\cite{ahn2022say,huang2022inner,huang2022language,wu2023tidybot,lin2023text2motion,huang2023voxposer,zhangboss,liu2023llm+p,dalal2024psl}. Work such as SayCan~\cite{ahn2022say} and TidyBot~\cite{wu2023tidybot} are similar to our own. They behavior clone / design a library of skills and use LLMs to perform task planning over the set of skills. Our work focuses primarily on designing the structure of skills for low-level control, decomposing them into motion planning and sim2real local policies. On the other hand, works such as LLMTrajGen~\cite{kwon2024language} and CoPa~\cite{huang2024copa} directly prompt VLMs to output sequences of end-effector poses, but are limited to short horizon tasks. Finally, PSL~\cite{dalal2024psl} and Boss~\cite{zhangboss} use LLMs to accelerate the RL training process for long-horizon tasks, yet must train on the test task, unlike our method which can solve a wide array of manipulation tasks zero-shot.

\noindent \textbf{Sim2real approaches in robotics}
Transferring RL policies trained with procedural scene generation enables generalist robot locomotion~\cite{lee2020learning,kumar2021rma,zhuang2023robot,agarwal2023legged,cheng2023parkour}. However, these policies are typically limited to a single skill, such as walking, or slight variations like different velocities or headings. 
Sim2real transfer has also been explored for dexterous manipulation skills~\cite{akkaya2019solving,andrychowicz2020learning,handa2022dextreme,agarwal2023dexterous,lum2024dextrah} and contact-rich manipulation~\cite{tang2023industreal,tang2024automate,jiang2024transic}. In our work, we train a variety of skills for manipulation and demonstrate zero-shot generalization on numerous unseen tasks. We outperform methods that use end-to-end sim2real transfer~\cite{peng2017simtoreal} as well as real world corrections~\cite{jiang2024transic}. \ours is orthogonal to human correction approaches, and can benefit from real-world data as well.
  
\vspace{-1pt}

\section{Methods}
\label{sec:methods}
To build agents capable of generalizing to a wide class of long-horizon robotic manipulation tasks, we propose a novel approach (\ours) that hierarchically decomposes manipulation tasks, takes advantage of the generalization capabilities of foundation models for vision and language and uses large-scale learning with our proposed policy class to learn manipulation skills. We begin by describing our framework (Fig.~\ref{fig:main fig}) and formulate local policies. We then discuss how to train local policies for sim2real transfer. 
Finally, we outline deployment: integrating VLMs, Motion Planning and sim2real policy learning to foster broad generalization. 

\vspace{-1pt}
\subsection{Framework}
We can decompose any task the robot needs to complete into a problem of learning a set of temporally abstracted actions (skills) as well as a policy over those skills~\cite{SUTTON1999181}. Given a language goal $g$, and observation $O$, we can select our high-level policy, $p_{\theta}(g_k | g, O)$ to be a pre-trained VLM, where $g_k$ is the $k$'th language subgoal. The choice of skill will be extracted from $g_k$ below. State-of-the-art VLMs have been shown to be capable of decomposing robotics tasks into high-level language subgoals~\cite{ahn2022say,huang2022inner,huang2022language,wu2023tidybot} because they are trained using a vast corpus of internet-scale data and have captured powerful, visually grounded semantic priors for what various real world tasks look like. 

\looseness=-1 Any policy class can be used to define the skills, denoted as $p_{\phi_k}(a^t|g_k, O^t)$, which take in the $k$th sub-goal $g_k$ and current observation $O^t$. However, note that many manipulation skills (\textit{e.g.} picking, pushing, turning, etc.) can be decomposed into a policy $\pi_{reach}$ to achieve target poses near objects, $X_{targ,k}$, followed by policy $\pi_{loc}$ for  contact-rich interaction. 
Accordingly, $p_{\phi_k}(a^t|g_k, O^t) = \pi_{reach}(\tau_{reach} | g_k, O^t) \pi_{loc}(a_{loc}^t | O_{loc}^t)$. To implement $\pi_{reach}$, we need to interpret language sub-goals $g_k$ to take the robot from its current configuration $q_{k,i}$ to some target configuration $q_{k,f}$ such that $X_{ee}$ (the end-effector pose) is close to $X_{targ,k}$. We structure the VLM's sub-goal predictions $g_k$ as (object, skill) tuples and interpret these plans into robot poses by pairing any language conditioned pose estimator or affordance model (to predict $X_{targ,k}$) with an inverse kinematics routine (to compute $q_{k,f}$). Motion planning is used to implement $\pi_{reach}$ by predicting collision-avoidant trajectories $\tau_{reach}$ to achieve the target configuration $q_{k,f}$. 

Finally, we instantiate local policies ($\pi_{loc}$) to be invariant to robot poses as well as object poses, order of skill execution and scene configurations with: 1) initialization region $s_{init}$ near a target region/object of interest which has pose $X_{targ,k}$, 2) local observations $O_{loc}^t$, independent of the absolute configuration of the robot and scene and only observing the environment around the interaction region and 3) actions $a_{loc}^t$relative to the local observations. Overall:
\begin{align*}
\pi_{loc}(a_{loc}^t | O_{loc}^t), s_{init} = \{s \mid ||X_{ee} - X_{targ,k} ||^2 < \epsilon\}
\end{align*}

\vspace{-1pt}
\subsection{Training Local Policies for Sim2Real Manipulation}

To train local policies, we adapt the standard two-phase training approach~\cite{agarwal2023dexterous,cheng2023parkour,zhuang2023robot,uppal2024spin,lum2024dextrah,jiang2024transic} in which we first train state-based expert policies using RL, then distill them into visuomotor policies for transfer. 
While local policies can generalize across scene arrangements, robot configurations, and object poses, broad object-level generalization requires diverse training data. To achieve this, we train a vast array of \textit{single-object} state-based RL policies and distill them into \textit{generalist} visuomotor policies per skill.
 
While such local policies can cover a broad set of manipulation skills (pick and place, articulated/deformable object manipulation, assembly, etc.), in this work, we focus on training the following skills $\pi_{loc}$: \textbf{pick}, \textbf{place}, \textbf{grasp handle}, \textbf{open} and \textbf{close} (Fig.~\ref{fig:sim vis}) as a minimal skill library to demonstrate generalist manipulation capabilities for a specific class of tasks. \textbf{Pick} grasps any free rigid objects. \textbf{Place} sets the object down near the initial pose. \textbf{Grasp Handle} grasps the handle of any door or drawer. \textbf{Open and Close} pull or push doors and drawers to open or close them. 

To train robust local policies via RL, they require a diverse set of training environments, carefully designed observations and action spaces and well-defined reward functions enabling them to acquire behaviors in a manner that will transfer to the real world. We describe how to in this section.
 
\noindent \textbf{Data Generation}
We need to first specify a set of objects to manipulate, an environment, and an initial local state distribution. 
For pick/place, we train on 3.5K objects from UnidexGrasp~\cite{xu2023unidexgrasp}, randomly spawned on a table top. To ensure local policies can learn obstacle avoidance and constrained manipulation, we spawn clutter objects and obstacles in the scene. We sample initial poses in a half-sphere, with the gripper pointing toward the object (for picking) and near the placement location (for placing). For local articulated object manipulation, the region of interaction only contains the handle (2.6K objects of Partnet~\cite{mo2019partnet}) and door/drawer surface (designed as cuboids). We randomize the size, shape, position, orientation, joint range, friction and damping coefficients, covering a wide set of real world articulated objects. We sample initial poses in a half-sphere around the handle (for grasp handle) and a randomly sampled initial joint pose (open/close). Finally we collect valid pre-grasp poses (antipodal sampling~\cite{sundermeyer2021contact}) for picking and grasping handles and rest poses (from UnidexGrasp) for learning placing.

\noindent \textbf{Observations} 
We use a single observation space for all RL experts, accelerating learning by incorporating significant amounts of privileged information.
Blind local policies can struggle to learn to manipulate objects with complex geometries as it is often necessary to have some notion of object shape to know how to manipulate. Thus, we propose to use a low-dimensional representation of the object shape by performing Farthest Point Sampling (FPS) on the object mesh with a small set number of desired key-points K (16). 
Furthermore, to ease the burden of credit assignment and thereby accelerate learning, we incorporate the individual reward components $\{\mathbf{r}\}$ and an indicator for the final observation $\mathds{1}\{t=T\}$.
RL observations are $O^t=\langle X_{ee}^t, \dot{X}_{ee}^t, X_{obj}^t, \{FPS_{obj}^t\}_{k=1}^K, \{\mathbf{r}\}^t, \mathds{1}\{t=T\}\rangle$

\noindent \textbf{Actions} We use the action space from Industreal~\cite{tang2023industreal} which has been shown to successfully transfer manipulation policies from sim2real for precise assembly tasks. Our policies predict delta pose targets for a Task Space Impedance (TSI) controller, where $a = [\Delta x; \Delta \theta]$, where $\Delta x$ is a position error and $\Delta \theta$ is a axis-angle orientation error. 

\noindent \textbf{Rewards} We train RL policies ($\pi_{loc_k}$) in simulation using reward functions we design to elicit the desired behavior per skill $k$.
We propose a reward framework that encompasses our local skills: $\mathbf{r} = c_{1}r_{ee} + c_{2}r_{obj} + c_{3}r_{ee,obj} + c_{4}r_{action} + c_{5}r_{succ}$. $\mathbf{r}$ specifies behavior for a broad range of manipulation tasks which involve moving the end-effector to specific poses (often right before contact) as well as a target object to desired poses and need to do so while maintaining certain constraints on the relative motion between the end-effector and the object as well as pruning out undesirable actions. 
$r_{ee}$ encourages reaching/maintaining specific end-effector poses, $r_{obj}$ restricts/encourages specific object poses or joint configurations, $r_{ee,obj}$ constrains the end-effector motion relative to the object(s) in the scene, $r_{action}$ restricts or penalizes undesirable actions and $r_{succ}$ is a binary success reward.  

\vspace{-1pt}
\subsection{Generalist Policies via Distillation}
In order to convert single-object, privileged policies into real world deployable skills, we distill them into multi-object, generalist visuomotor policies using DAgger~\cite{ross2011reduction}.

\noindent \textbf{Multitask Online Imitation Learning}
Empirically the standard, off-policy version of DAgger with interleaved behavior cloning (to convergence) and large dataset collection does not perform well. The policy ends up modeling data from policies whose state visitation distributions deviate significantly from the current policy. On the other hand, on-policy variants of DAgger, which take a single gradient step per environment step~\cite{agarwal2023legged,agarwal2023dexterous,uppal2024spin,lum2024dextrah}, can produce unstable results in the multi-task regime since the policy only gets data from a single object in a batch. We introduce a simple variant of DAgger which smoothly trades off between the two extremes by incorporating a replay buffer of size $K$ that holds the last $K*B$ trajectories in memory.  Training alternates between updating the agent for a single epoch on this buffer and collecting a batched set of trajectories (size $B$) from the environment for the current object. 

\noindent \textbf{Observation Space Design for Locality}
For local policies to transfer effectively to the real robot, the observation space and augmentations must be designed with transfer in mind. To imitate a privileged expert, our observation space must be expressive -- providing as much information as possible to the agent. The observations must also be local to enable all of the properties of locality, and augmentations must ensure the policy is robust to noisy real world vision.

Local observations use wrist camera depth maps. Depth maps transfer well from sim2real for locomotion~\cite{agarwal2023legged,zhuang2023robot,cheng2023parkour,uppal2024spin}, and wrist views are inherently local and improve manipulation performance~\cite{hsu2022vision,dalal2023optimus,robomimic2021}.
To further enforce locality, we clamp depth values to a max depth of $30cm$ and then normalize the values to between $0$ and $1$. 
Since local wrist-views often get extremely close to the object during execution, it can become difficult for the agent to understand the overall object shape. Thus, we include the initial local observation $O_{loc,depth}^0$ at every step with a segmentation mask of the target object ($O_{loc,seg}^0$) so that the local policy is aware of which object to manipulate.
We transform absolute proprioception into local by computing observations relative to the first time-step ($O_{loc,ee} = [X_{ee,t}^0-X_{ee}^0]$) and incorporate velocity information ($\dot{O_{loc,ee,t}}$), which improves transfer.
Our observation space is $\mathbf{O_{loc}^t} = \langle O_{loc,depth}^t, O_{loc,seg}^0, O_{loc,depth}^0, O_{loc,ee}^t, \dot{O_{loc,ee}^t} \rangle$.

\noindent \textbf{Augmentations} To enable robustness to noisy real world observations, namely edge artifacts and irregular holes, we augment the clean depth maps we obtain in simulation. For edge artifacts, in which we observe dropped pixels and noisiness along edges, we use the correlated depth noise via bi-linear interpolation of shifted depth from ~\cite{barron2013intrinsic} which tends to model this effect well. We also observe that real world depth maps tend to have randomly placed irregular holes (pixels with depth 0). As a result, we compute random pixel-level masks and Gaussian blur them to obtain irregularly shaped masks that we then apply to the depth image.  We also use random camera cropping augmentations which has been shown to improve visuomotor learning performance~\cite{robomimic2021}. 

\vspace{-1pt}
\subsection{Zero-shot Text Conditioned Manipulation}
Given our framework and trained local policies, how do we now deploy them in the real-world, to solve a wide array of manipulation tasks in a zero shot manner? 

To enable our system to solve long-horizon tasks, $p_{\theta}(g_k | g, O)$, decomposes the task into a skill chain to execute given task prompt $g$. We implement $p_{\theta}$ as GPT-4o, a SOTA VLM. Given the task prompt $g$, descriptions of the pre-trained local skills and how they operate, and images of the scene $O$, we prompt GPT-4o to give a plan for the task structured as a list of (object, skill) tuples. For example, for the task shown in Fig.~\ref{fig:main fig}, GPT outputs ((handle, open), (rice pick), (microwave, place), (handle, close)). We then need a language conditioned pose estimator (to compute $X_{targ,k}$) that generalizes broadly; we opt to use Grounded SAM~\cite{ren2024grounded} due to its strong open-set segmentation capabilities. To estimate $X_{targ,k}$, we can segment the object pointcloud, average it to get a position and use its surface normals to select a collision-free orientation. One issue is that Grounding Dino~\cite{liu2023grounding}, used in Grounded SAM, is very sensitive to the prompt. As a result, we pass its predictions back into GPT-4o to adjust the object prompts to capture the correct object. 

For predicting $\tau_{reach}$, while any motion planner can be used, we select Neural MP~\cite{dalal2024neuralmp} for its fast (2s) and strong real-world planning. Given $X_{targ,k}$, we compute target joint state $q_{k,f}$, plan with Neural MP open-loop and execute the predicted $\tau_{reach}$ on the robot using a PID joint controller. We then execute the appropriate local policy (as predicted by the VLM) on the robot to perform manipulation. We alternate between motion planning and deploying local policies until the task is complete. Finally, we note that the particular choice of models is orthogonal to our method.

\section{Experimental Setup and Results}
\label{sec:setup}

We pose the following experimental questions that guide our evaluation: 1) Can an autonomous agent control a robot to perform a wide array of \textit{long-horizon} manipulation tasks zero-shot?
2) How does our approach compare to methods that learn from online interaction? 
3) For direct sim2real transfer, how do Local Policies compare against end-to-end learning and other transfer techniques that leverage human correction data?
4) To what degree do the design decisions made in \ours affect the performance of the method?

\vspace{-1pt}
\subsection{Training and Deployment Details}
\noindent \textbf{Architecture and Training}
We train all RL policies at scale using PPO~\cite{schulman2017proximal} in GPU-parallelized simulation~\cite{makoviychuk2021isaac}. We train for 500 epochs, with an environment batch size of 8192 and max episode length of 120 steps per skill. To learn visuomotor policies to perform high-frequency (60 Hz) end-effector control, we pair Resnet-18~\cite{he2024learning} and Spatial Soft-max~\cite{finn2016deep} with a two layer MLP decoder (4096 hidden units). 
Finally, for training, minimizing Mean Squared Error loss is sufficient for learning multitask policies via DAgger. In early experiments, we found that our architecture performs comparably to using LSTMs~\cite{hochreiter1997long}, Transformers~\cite{vaswani2017attention}, and ACT~\cite{zhao2023learning} and is faster to train (5-10x) and deploy (2x). 

\looseness=-1 \noindent \textbf{Hardware Setup}
We use the Franka Panda robot arm with the UMI~\cite{chi2024universal} gripper fingertips and a wrist-mounted Intel Realsense d405 camera for obtaining local observations (84x84 resolution). 
Note for local observations, using depth sensing that is accurate at short range (such as the d405) is crucial to obtaining high quality local depth maps. 
We perform hole-filling and smoothing to clean the depth maps. 
Following Transic~\cite{jiang2024transic}, we do not model the compliance of the UMI gripper in simulation, but instead transfer policies trained with rigid fingertips to the real world, which performs well in practice. 
For real world control, we use a TSI end-effector controller at 60 Hz with (Leaky) Policy Level Action Integration (PLAI)~\cite{tang2023industreal}. We use Leaky PLAI with $.001$ position action scale, $.05$ rotation action scale for pick and $.005$ rotation action scale for all other skills. Finally, we use 4 calibrated Intel Realsense d455 cameras for global view observations (640x480). 

\vspace{-1pt}
\subsection{Simulation Comparisons and Analysis}
\noindent \textbf{Robosuite Benchmark Results} We first evaluate against the long-horizon manipulation tasks used in PSL~\cite{dalal2024psl} from the Robosuite benchmark~\cite{zhu2020robosuite} in simulation which has a set of challenging long-horizon manipulation tasks (\textbf{PickPlace\{Bread, Milk, Cereal, CanBread, CerealMilk\}}). 
We compare to end-to-end RL methods~\cite{yarats2021mastering}, hierarchical RL~\cite{dalal2021accelerating,dalal2024psl}, task and motion planning~\cite{garrett2020pddlstream} and LLM planning~\cite{ahn2022say}. 
In these experiments, we \textit{zero-shot} transfer our trained policies to Robosuite and evaluate their performance against methods that use task specific data (Tab.~\ref{table:psl results}). \ours outperforms or matches PSL, the SOTA method on these tasks, across the board, achieving an average success rate of $97.33\%$ compared to $95.83\%$. These results demonstrate that \ours can outperform methods that are trained on the task of interest~\cite{dalal2024psl,dalal2021accelerating,yarats2021mastering} as well as planning methods that have access to privileged state info~\cite{garrett2020pddlstream,ahn2022say}.
\begin{table}[t]
\vspace{5pt}
\resizebox{\linewidth}{!}{%
\begin{tabular}{cccccccc}
\toprule
\multicolumn{1}{l}{} & Bread      & Can        & Milk       & Cereal     & CanBread   & CerealMilk & Average \\
\midrule
\textit{Stages}      & \textit{2} & \textit{2} & \textit{2} & \textit{2} & \textit{4} & \textit{4} & \\
\midrule
\textit{Online Learning:} \\
DRQ-v2                  & $52\%        $ & $32  \%    $   & $2  \%    $    & $0   \%     $  &$ 0\%   $      & $0  \%   $ &  14\%\\
RAPS                 & $0 \%       $  & $0    \%  $    & $0   \%  $     & $0    \%   $   & $0  \%  $      &$ 0  \%  $     & 0\% \\
PSL                 & $100  \% $     & $100   \% $    & $100   \%  $   & $100   \%   $  & $90   \%  $    & $85   \%   $ &  96\% \\
\midrule
\textit{Zero-Shot:} \\
TAMP                 & $90 \%     $   & $100  \%   $   & $85   \%   $   & $100  \%    $  & $72  \%  $     & $71  \%   $   & 86\% \\
SayCan               & $93  \%   $    & $100   \%  $   & $90   \%   $   & $63   \%   $   & $63   \%  $    & $73   \%  $   & 80\% \\
\textbf{Ours}               & $100  \%$      & $100   \%  $   & $99    \%  $   & $97    \%  $   & $97   \%  $    & $91   \%   $  & \textbf{97\%} \\
\bottomrule
\end{tabular}
}
\vspace{-5pt}
\caption{
\small \textbf{Robosuite Benchmark Results.} \ours zero-shot transfers to Robosuite, outperforming end-to-end and hierarchical RL methods as well as traditional and LLM planning methods.}
\label{table:psl results}
\vspace{-17pt}
\end{table}

\noindent \textbf{\ours Analysis and Ablations}.
We study design decisions proposed in our method by training single object pick policies on 5 objects (remote, can, bowl, bottle, camera) and testing on held out poses. We begin with our observation space design choices: \ours achieves $97.44\%$ success rate in comparison to (94.33\%, 96.64\%, 97.25\%) for removing key-point observations, success observation and reward observations respectively. Incorporating key-point observations is the most impactful change, enabling the agent to perceive the shape of the target object. Next, we evaluate how the level of locality (the size of the region around the target object that we initialize over) affects learning performance. At convergence, we find that \ours (8cm max distance from target) achieves $97.44\%$ success rate while performance diminishes with increasing distance ($95.65\%$, $89.55\%$, $72.52\%$) for $16$cm, $32$cm and $64$cm respectively. 

For DAgger, we analyze our observation design choices and find that including velocity information, the first observation, and changing proprioception to be relative to the first frame are crucial to the success of our method. While \ours gets $94.3\%$ success, removing velocity info and using absolute proprioception hurt significantly ($89.92\%$ and $90.94\%$) while removing the first observation drops performance to $93.13\%$. We also vary the DAgger buffer size, from $1$ (on-policy), $10$, $100$, and $1000$ (off-policy) for multitask training (with 3.5K objects, not 5). We find that 100 performs best, achieving 85\% in simulation averaged across 100 held out objects, out performing (78\%, 82\% and 75\%) for 1, 10 and 1000 respectively. 

\vspace{-1pt}
\subsection{Real World Evaluation}
\noindent \textbf{FurnitureBench Results}
To evaluate the sim2real capabilities of local policies (Tab.~\ref{table:transic results}), we deploy \ours on  FurnitureBench~\cite{heo2023furniturebench}, comparing against a wide array of direct-transfer~\cite{peng2017simtoreal}, imitation learning~\cite{NIPS1988_812b4ba2, mandlekar2021matters}, offline RL~\cite{kostrikov2021offline} and human-in-the-loop methods~\cite{jiang2024transic, kelly2018hgdagger, mandlekar2020humanintheloop} from Transic~\cite{jiang2024transic}. These tasks are single stage; we train local policies to perform pushing (\textbf{Stabilize}), picking (\textbf{Reach and Grasp}) and insertion (\textbf{Insert}). We predict a start pose to initialize the local policy from and deploy the simulation-trained policies. \ours matches or outperforms end-to-end direct transfer methods ($75\%$, $53.3\%$), imitation methods ($55\%$, $82.7\%$, $65\%$, $75\%$, $86.7\%$) and sim2real methods that leverage additional correction data~\cite{jiang2024transic}. 
For Insert, local policies are able to outperform Transic without using any real world data, achieving 80\% while Transic achieves 45\%. These experiments demonstrate \ours improves over end-to-end learning and is capable of handling challenging initial states, contact-rich interaction and precise motions.

\begin{table}
\centering
\vspace{2pt}
\resizebox{\linewidth}{!}{%
\begin{tabular}{@{}cccccccc@{}}
\toprule
Tasks  & Ours  & Transic & \begin{tabular}[c]{@{}c@{}}Direct\\ Transfer\end{tabular} & \begin{tabular}[c]{@{}c@{}}DR. \& Data\\ Aug.~\cite{peng2017simtoreal}\end{tabular} & \begin{tabular}[c]{@{}c@{}}HG-Dagger~\cite{kelly2018hgdagger}\end{tabular} & \begin{tabular}[c]{@{}c@{}}IWR~\cite{mandlekar2020humanintheloop}\end{tabular} & BC~\cite{NIPS1988_812b4ba2} \\ \midrule
Stabilize       & $95\%$        & \textbf{100\%} & 10\%           & 35\%            & 65\%           & 65\%       & 40\% \\
Reach and Grasp & \textbf{95\%} & \textbf{95\%}  & 35\%           & 60\%            & 30\%           & 40\%       & 25\% \\
Insert          & \textbf{80\%} & 45\%           & 0\%            & 15\%            & 35\%           & 40\%       & 10\% \\
\midrule
Avg             & \textbf{90\%} & 80\%           & 15\%           & 36.7\%          & 43.3\%         & 48.3\%     & 25\% \\
\bottomrule
\end{tabular}
}
\vspace{-5pt}
\caption{
\small \textbf{Transic Benchmark Results} \ours achieves SOTA results on the Transic~\cite{jiang2024transic} benchmark in terms of task success rate without using any real world data, outperforming direct transfer, imitation learning and human-in-the-loop methods.
}
\label{table:transic results}
\vspace{-18pt}
\end{table}

\noindent \textbf{Zero-shot Long-horizon Manipulation}
To test the generalization capabilities of our method, we propose 5 diverse long-horizon manipulation tasks (Fig.~\ref{fig:teaser}) which involve pick and place, obstacle avoidance and articulated object manipulation. \textbf{Cook}: put food into a pot on a stove (2 stages), \textbf{Replace}: take a pantry item out of the shelf, put it on a tray and take an object from the tray and put it in the shelf (4 stages), \textbf{CabinetStore}: open a drawer in the cabinet, put an object inside and close it (4 stages). \textbf{DrawerStore}: open a drawer, put two personal care items inside and close the drawer (6 stages) and \textbf{Tidy}: clean up the table by putting all the toys into a bin (8 stages). Each task has a unique object set (5 objects), receptacle (pot, shelf, etc.) and text description. We run 10 evaluations per task, randomizing which objects are present and their poses, receptacle poses, and target poses. All poses are randomized over the table and we select a diverse set of evaluation objects. 

\noindent \textbf{Evaluation Criteria} For each task we identify the stages required for completion. A trial is considered successful if the final state meets the task's goal as specified. Additionally, we track the number of stages completed in each trial. We conduct 10 trials per task, reporting the success rate and average number of stages completed.

\looseness=-1 \noindent \textbf{Comparisons}
We evaluate SOTA text-conditioned manipulation approaches: SayCan~\cite{ahn2022say}, LLMTrajGen~\cite{kwon2024language} and VoxPoser~\cite{huang2023voxposer}. For SayCan, we use our VLM and motion planning system with engineered interaction primitives to assess the role of training local policies. We additionally compare against a pre-trained manipulation model OpenVLA~\cite{kim2024openvla}, fine-tuning it per task using 25 demonstrations on held-out objects, poses, and scene configurations. Given a text prompt, we record task success rates and stages completed.

\begin{table}[t]
\vspace{2pt}
\centering
\resizebox{\linewidth}{!}{%
\begin{tabular}{lcccccc}
\toprule
 & Cook & Replace & CabinetStore & DrawerStore  & Tidy & Avg \\
\midrule
\textit{Stages}      & \textit{2} & \textit{4} & \textit{4} & \textit{6} & \textit{8} & \textit{4.8} \\
\midrule
OpenVLA              & 0\% (0.1)      & 0  (0.0)     & 0\% (0.0)      & 0 (0.0)     & 0  (0.0)    & 0\% (.02)   \\
SayCan               & 80\%  (1.7)    & 10\% (1.3)      & 70\% (3.5)      & 20\%  (3.6)    & 20\% (4.8)      & 40\% (3.0)   \\
LLMTrajGen        & 70\%   (1.5)   & 0\% (0.6)      & 0\% (0.6)      & 0\% (1.0)     & 0\% (2.6)      & 14\% (1.3)   \\
VoxPoser            & 70\% (1.4)      & 0\% (0.8)      & 0\% (0.8)      & 0\% (0.9)      & 10\% (4.4)      & 16\% (1.7)   \\
\textbf{Ours}                 & \textbf{90\% (1.9)}      & \textbf{80\% (3.7)}      & \textbf{90\% (3.9)}      & \textbf{60\% (4.7)}      & \textbf{60\% (7.2)}      & \textbf{76\% (4.3)}   \\
\bottomrule
\end{tabular}
}
\vspace{-5pt}
\caption{\small \textbf{Zero-shot Long Horizon Manipulation} We report task success rate and average number of stages completed per real world task. \ours outperforms all methods on each task, achieving 76\% with 4.28/4.8 stages completed on average.}
\label{table:main results}
\vspace{-18pt}
\end{table}

\looseness=-1 Across all 5 tasks (Tab.~\ref{table:main results}), \ours achieves \textbf{$76\%$ zero-shot success rate}, outperforming all methods. Notably, our local policies are not trained on \textit{any of these specific objects} or in \textit{these specific configurations} and require \textit{no real-world adaptation}. \ours is able to avoid obstacles while performing manipulation of unseen objects in arbitrary poses and configurations. Failure cases stemmed from 1) vision failures as open-set detection models such as Grounding Dino~\cite{liu2023grounding} detected the wrong object, 2) imperfect motion planning, resulting in collisions with the environment during execution which dropped objects sometimes and 3) local policies failing to manipulate from sub-optimal initial poses. In general, DrawerStore and Tidy are the most challenging tasks due to their horizon, and consequently all methods, including our own perform worse (60\% for ours, 20\% for best baseline). 

SayCan is the strongest baseline (40\% success), achieving non-zero success on every task by leveraging the generalization capabilities of vision-language foundation models in a structured manner. However, when initial poses are not ideal or the task requires contact-rich control, pre-defined primitives fall apart (10-20\% success).
LLMTrajGen excelled at top-down pick-and-place (Cook: 70\%) but struggled with obstacle avoidance (Replace) and articulated object manipulation (Store) due to limited prompt coverage.
VoxPoser matches LLMTrajGen in Cook but failed at precise rotations and tight-space tasks (Replace, Store) and frequently generated incorrect plans for long-horizon tasks (Tidy).
Finally, OpenVLA failed to solve any task, unable to generalize to held-out objects and poses even with few-shot data. We attempted to evaluate it on its training objects and it still performs poorly with strong pose randomization. 

\vspace{-1pt}
\section{Discussion}
\vspace{-1pt}
\label{sec:conclusion}
\looseness=-1 We present \ours, a method for  long-horizon manipulation tasks with unseen objects and configurations via generalist policies for sim2real transfer. 
We propose local policies, a novel policy class invariant to pose, skill order, and scene configuration, enabling broad generalization. For deployment, we leverage foundation models for vision, language and motion planning to solve long-horizon manipulation tasks from text prompts. 
Across 50 real-world long-horizon manipulation tasks, \ours achieves 76\% \textit{zero-shot} success, surpassing SOTA planning and imitation methods.

\newpage

\section*{Acknowledgement}
We thank Russell Mendonca, Ananye Agarwal and Yash Narang for their insightful discussions and feedback. We additionally thank Russell Mendonca, Shikhar Bahl, Lili Chen, and Unnat Jain for feedback on early drafts of this paper. This work was supported in part by the NSF Graduate Fellowship, Apple, and ONR grant N00014-23-1-2368.

\bibliographystyle{IEEEtran}
\bibliography{main}
\input{supplement}

\end{document}

%% file: supplement.tex
\clearpage

\section*{Appendix}

\section{RL Training Details}
In this section, we provide a detailed description of the data generation process, the exact reward definitions and the specific hyper-parameters we use to train our skills.

\subsection{Data Generation}

For training generalist pick and place skills, we require a large dataset of common objects that can simulate well with contact. As a result, we train policies using the UnidexGrasp dataset~\cite{xu2023unidexgrasp} which contains 3.5K objects of 133 categories such as bowls, cups, bottles, cameras, remotes, etc. Since our policies are local, we can simply generate scenes with a single object spawned on a table top. However, such an agent may not generalize well to manipulation in tight spaces and among clutter, when it needs to perform local obstacle avoidance and constrained manipulation. For robustness, we train with randomly sampled clutter objects (UnidexGrasp) and obstacles (cuboids) that we spawn in the scene. 

For picking, to define an initial state distributions which ensures locality (within $\epsilon$ of the target object), we sample poses in a half-sphere above the table with radius $\epsilon=0.08$ and an additional error tolerance of 0.05 around the target object that are always pointing towards the object (ensuring the object is visible from the wrist camera). For placing, we execute the pick policy and then sample initial poses in a cuboid of with side length $\epsilon=0.2$ around the picking pose. 
In both cases we ensure that the sampled poses are not in contact with anything in the environment (aside from the in-hand object if present).

For local articulated object manipulation of objects such as doors and drawers, the design and global structure of the asset is not important. In fact, the only component of interest to the local policy is the handle. Accordingly, we sample from a dataset of 2.6K door and drawer handles from the PartNet dataset~\cite{mo2019partnet} and build door and drawer assets out of cuboids (Fig.~\ref{fig:sim vis}) as they are straightforward to randomize. We define drawers as boxes to be pulled straight out and doors as boxes to be opened using a vertical hinge joint. We randomize the size, shape, position, orientation, articulated joint range, friction and damping coefficients of the articulated objects, which covers a wide set of real world articulated objects. Detailed randomization distributions are presented in Tab.~\ref{tab:randomization_details}. For the grasp handle skill, we sample initial poses pointing toward the handle in a half sphere (in this case vertical half-sphere, away from the door) with radius $\epsilon=0.08$ and an additional error tolerance of 0.05. For opening and closing, after sampling a random initial pose of the articulated joint, we execute the grasp handle policy and add a small noise to ensure diversity of the final end-effector pose.

Finally, we collect valid pre-grasp and rest poses in simulation to help train our local policies. Specifically, we randomly sample grasp poses on the object mesh using antipodal sampling~\cite{sundermeyer2021contact} (1K per rest pose for UniDexGrasp objects and 2.5K for PartNet objects). We then move the Franka arm to pick/grasp the handle of the object using the pre-sampled grasp poses and save the successful poses. We also utilize the success rate of this scheme to filter out rest object poses that are not graspable (\textit{e.g.}, an upside down bowl). We also  generate a wide set of object rest poses for training the placing policy using the initial poses from the UnidexGrasp dataset, and augmenting them by rotating about the z-axis with 8 different angles and testing to ensure the objects remain at rest.

\begin{table}[H]
    \centering
    \begin{tabular}{c|c}
    \toprule[1.2pt]
       Parameter & Range / Distribution \\
       \midrule
        \multicolumn{2}{c}{\textit{UniDexGrasp objects}} \\
        Object size & $[0.06, 0.30]$\\
        Initial object position (XY) & $\mathcal{U}(0.3, 0.7) \times \mathcal{U}(-0.2, 0.2)$\\
        Initial object rotation (Z-axis) & $\mathcal{U}(-\pi, \pi)$ \\
        \midrule
        \multicolumn{2}{c}{\textit{Articulated objects}} \\
        Door size & $\mathcal{U}(0.25, 0.40) \times \mathcal{U}(0.20, 0.50)$\\
        Door damping & $\mathcal{U}(0.01, 0.02)$\\
        Door friction & $\mathcal{U}(0.025, 0.050)$\\
        Door joint range & $[0, \pi/2 ]$ \\
        Drawer size & $\mathcal{U}(0.25, 0.50) \times \mathcal{U}(0.08, 0.25)$\\
        Drawer damping & $\mathcal{U}(0.10, 0.20)$\\
        Drawer friction & $\mathcal{U}(0.25, 0.50)$\\
        Drawer joint range & $[0, 0.3]$ \\
        Distance to robot base & $\mathcal{U}(0.65, 0.75)$\\
        Object Orientation & $\mathcal{U}(-\pi/2, \pi/2)$\\
    \bottomrule[1.2pt]
    \end{tabular}
    \caption{Randomization for data generation.}
    \label{tab:randomization_details}
\end{table}

\subsection{Rewards}
We provide additional details on how to specify rewards for each skill. Specifically, we define $r_{ee}, r_{obj}, r_{ee,obj}$ and $r_{action}$.

\noindent \textbf{Pick} involves moving the gripper so the object can be easily grasped. 
Instead of encouraging the agent to move towards the overall object pose, which is not necessarily the pose to achieve for grasping, we provide dense signal for learning to grasp using pre-sampled grasp poses ($\{X_{targ}\}$). We encourage the agent to minimize the key-point distance~\cite{allshire2022transferring} to the nearest grasp pose:
\begin{align*}
    r_{ee,grasp} = \sum_{i=1}^Ne^{\min_{\{X_{targ}\}}||X_{ee}^i-X_{targ}^i||^2}
\end{align*}

Another challenge is that of picking in tight spaces, in which the policy changing directions and too frequently may cause damage and failure to complete the task. Thus, we encourage the agent to minimize its change in gripper orientation while interacting, by adding a penalty term on the angle between the current and previous gripper pose along the gripper's central axis ($\varv$), 
\begin{align*}
r_{ee, init} = -\arccos((R_{ee}^t\varv)(R_{ee}^{t-1}\varv))
\end{align*}
We also add a term to minimize the contact force on the gripper which discourages contact with any part of the scene. 
\begin{align*}
    r_{ee,obj} = \max (\max(f_{left}, f_{right}), 0)
\end{align*}
Finally, when picking, the agent should try to minimize moving the target object: we set 
\begin{align*}
r_{obj} = e^{||X_{obj,xy}^t - X_{obj,xy}^0||^2}
\end{align*}
 to penalize changes in object pose. All other reward components have a constant set to 0.
 
\noindent \textbf{Place} requires the agent to carefully set the object down near the initial pose. 
To provide dense signal for placing and ensure stability, we use a key-point distance reward on the object pose to encourage it to reach a stable absolute rest pose ($r_{obj}$) 
\begin{align*}
    r_{obj} = e^{\sum_{i=1}^{8}||X_{obj,FPS}^i - X_{rest,FPS}^i||^2}
\end{align*}
and a key-point distance reward to the nearest grasp pose to encourage the agent to maintain a stable pose relative to the object itself while moving ($r_{ee}$). This reward is the same as in pick.

\noindent \textbf{Grasp Handle} enables the agent to be able to grasp the handle of any door or drawer, a necessary skill to interact with articulated objects. This skill uses similar rewards to pick, but adapted for articulation: 1) key-point-based rewards for reaching pre-grasp poses ($r_{ee}$) 2) restricting the gripper orientation in task irrelevant directions ($r_{ee, z}$) 3) restricting the gripper orientation relative to the handle in the x-axis ($r_{ee,obj}$) 4) discouraging the agent from moving the object by minimizing the motion of the joint ($r_{obj}$).

Same as in pick, we wish to use pre-grasp poses in order to provide dense signal to the agent and ensure that it does not change orientation too much. We use the same key-point-based reward $r_{ee,grasp}$, but instead restrict the policy's movement in the z-direction, an axis along which motion is not beneficial to solving the task 
\begin{align*}
    r_{ee,z} = -|X_{ee,z}^t-X_{ee,z}^0|
\end{align*}
We further encourage the agent to only move in task relevant directions, by encouraging the agent to minimize the angle between the gripper and handle, in the x-axis of the door frame:
\begin{align*}
    r_{ee,obj} = e^{\min((R_{handle}^tR_{ee}^t\varv)_x + thresh, 0)}
\end{align*}
Similar to pick, we discourage the agent from moving the object, in this case by minimizing the motion of the joint 
\begin{align*}
 r_{obj} = -|q^t - q^0|
\end{align*}

\noindent \textbf{Open and Close} involve opening and closing articulated objects, having already grasped the handle. We can solve the task with a dense absolute difference reward ($r_{obj}$) between the current and target joint angles (0 for close, and $q_{limit}$, the joint limit of the object for open). To ensure the agent maintains its grasp while smoothly moving the articulated to joint to the desired configuration, we use $r_{ee,z}$ from Grasp handle, penalize movement relative to the door/drawer handle ($r_{ee,obj}$), and discourage ($r_{action}$) taking actions that cause the handle to slip out (moving in the y or z axes).

These skills utilize a dense reward for solving the task 
\begin{align*}
    r_{obj} = |q^t - q_{targ}|
\end{align*} where $q_{targ}$ is 0 for close and is $q_{limit}$ the joint limit of the object for open. As with grasp handle, we restrict the policy's movement along the z-axis using $r_{ee,z}$. We additionally penalize any movement of the end-effector in the handle frame, 
\begin{align*}
    r_{ee,obj} = ||R_{handle}^tX_{ee}^t - R_{handle}^{t-1}X_{ee}^{t-1}||
\end{align*} and any sampled actions that cause the agent to move in the y or z axes: 
\begin{align*}
    r_{action} = -||(R_{handle}^t(a))_{yz}||^2
\end{align*}
These rewards ensure that the agent maintains its grasp while smoothly moving the articulated to joint to the desired configuration.

We train all RL policies with PPO~\cite{schulman2017proximal}. Detailed hyper-parameters are presented in Tab.~\ref{tab:rl_training_details} and Tab.~\ref{tab:rl_training_epochs}.
\begin{table}[H]
    \centering
    \begin{tabular}{c|c}
    \toprule[1.2pt]
       Hyperparameter & Value \\\hline
        Num. envs (Isaac Gym, state-based) & 8192\\
        Num. rollout steps per policy update & 32\\
        Num. learning epochs & 5\\
        Episode length & 120\\
        Discount factor & 0.99\\
        GAE parameter & 0.95\\
        Entropy coeff. & 0.0\\
        PPO clip range & 0.2\\
        Learning rate & 0.0005\\
        KL threshould for adaptive schedule & 0.16 \\
        Value loss coeff. & 4.0\\
        Max gradient norm & 1.0 \\
    \bottomrule[1.2pt]
    \end{tabular}
    \caption{Hyper-parameters for PPO.}
    \label{tab:rl_training_details}
\end{table}

\begin{table}[h]
    \centering
    \begin{tabular}{c|ccc}
    \toprule[1.2pt]
       Skill & \#Max epoch & \#Save best & \#Early stop  \\\hline
        Pick & 500 & 100 & 200 \\
        Place & 500 & 100 & 200 \\
        Grasp Handle & 500 & 100 & 100 \\
        Open & 500 & 100 & 100 \\
        Close & 500 & 100 & 100 \\
    \bottomrule[1.2pt]
    \end{tabular}
    \caption{Training epochs and early stop criterion for each task. \#Max epoch is the maximum number of iterations to run. \#Save best is the first iteration to begin saving best checkpoints. \# Early stop suggests terminating early if there is no improvement after certain number of iterations.}
    \label{tab:rl_training_epochs}
\end{table}

\section{Distillation Training Details}

\subsection{Multitask DAgger}
In our multitask DAgger implementation, we incorporate a replay buffer of size $K$ that holds the last $K\times B$ trajectories in memory. The training process alternates between updating the policy for a single epoch on this buffer and collecting a batched set of trajectories (size $B$) from the environment for the current object. 
In practice, we find that K=100, B=32 performs well, which means for a single object we collect 32 simultaneous trajectories at a time, and we can hold data for up to 100 objects in our buffer which is constantly refreshed as we collect new data. A practical issue is loading all objects from the dataset into simulation simultaneously is unfeasible. To address this, we split the dataset into batches of 100 objects and sequentially launch training on each batch for 100 epochs. Detailed multitask DAgger parameters are presented in Tab.~\ref{tab:dagger_training_details}.

\begin{table}[H]
    \centering
    \begin{tabular}{c|c}
    \toprule[1.2pt]
       Hyperparameter & Value \\\hline
        Num. envs (Isaac Gym, vision) & 128\\
        Episode length & 120\\
        Num. rollout steps per policy epoch & 120\\
        Num. learning epochs & 1\\
        Buffer size & 100 * 128 \\
        Learning rate & 0.0001\\
        Batch size & 2048 \\
    \bottomrule[1.2pt]
    \end{tabular}
    \caption{Hyper-parameters for Multitask DAgger.}
    \label{tab:dagger_training_details}
\end{table}

\subsection{Data Augmentation}

In addition to random camera cropping, we also apply \textit{edge noise} and \textit{random holes} to enhance robustness to real world observations.

\textbf{Edge artifacts} To model the noisiness along object edges, we use the correlated depth noise via bi-linear interpolation of shifted depth.  Given a depth map of size $H\times W$, we construct a grid $\{0,\cdots, H-1\}\times \{0,\cdots, W-1\}$. For each node on the grid, we apply a random shift $\mathcal{N}(0, 0.5)$ with probability $0.8$. We then perform bilinear interpolation between the original depth values and the adjusted grid to generate a new depth map.

\textbf{Random holes} We observe that, even after hole filling, the real world depth maps still contain irregular holes (especially for reflective surfaces and dim environments). To model these holes, we create a random pixel-level mask from $\mathcal U(0, 1)$. This mask is smoothed with Gaussian blur and normalized to the range $[0, 1]$. Based on the mask, we zero out pixels with mask values exceeding a threshold randomly sampled from $\mathcal U(0.6, 0.9)$. The randomization is applied to a depth map with probability $0.5$.

\begin{figure}[ht!]
    \centering
    \includegraphics[width=1.0\linewidth]{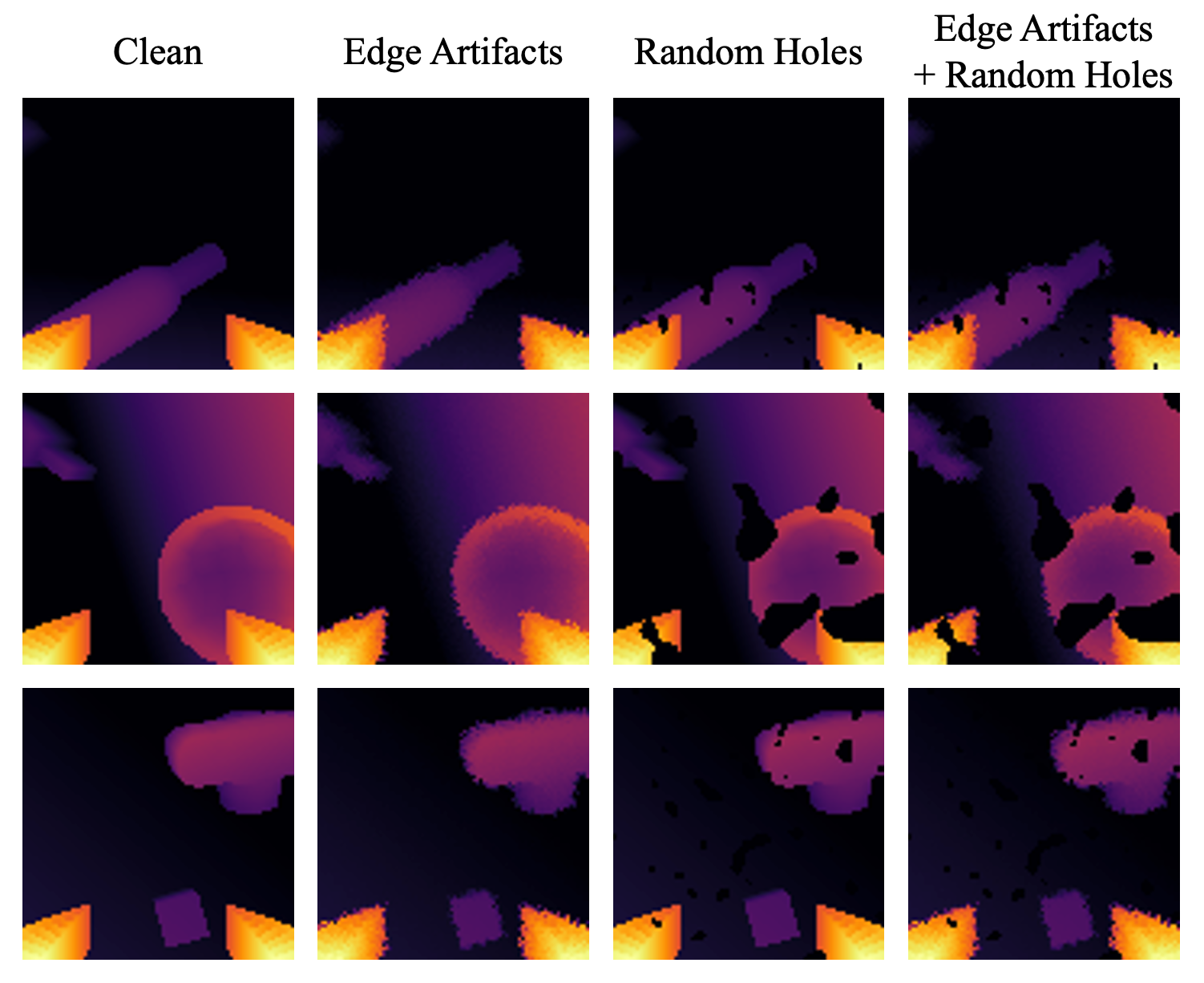}
    \caption{\textbf{Depth Augmentation} Visualization of edge artifacts and random holes on depth maps.}
    \label{fig:vis_depth_aug}
\end{figure}

We summarize hyper-parameters for DAgger data augmentation in Tab.~\ref{tab:dagger data aug} and visualize them in Fig.~\ref{fig:vis_depth_aug}. Note that the resolution of our depth map is $84\times 84$. The visual effect with these hyper-parameters will change on different resolutions.

\begin{table}[H]
    \centering
    \begin{tabular}{c|c}
    \toprule[1.2pt]
       Hyper-parameter & Value \\\hline
       \textit{Edge Noise} & \\
       Gaussian Noise Std & 0.5 \\
       Noise Accept Prob & 0.8 \\\hline
       \textit{Random Holes} & \\
       Gaussian Blur Kernel Size & [3, 27] \\
       Gaussian Blur Std & [1, 7] \\
       Mask Threshold & [0.7, 0.9] \\
       Hole Keep Prob & 0.5 \\
    \bottomrule[1.2pt]
    \end{tabular}
    \caption{Hyper-parameters for DAgger data augmentation.}
    \label{tab:dagger data aug}
\end{table}

\section{Deployment Details}
In this section, we describe our real-world deployment system in detail. 
We begin by providing a high-level overview of ManipGen deployment in pseudocode below:

\begin{figure}
    \centering
    \includegraphics[width=\linewidth]{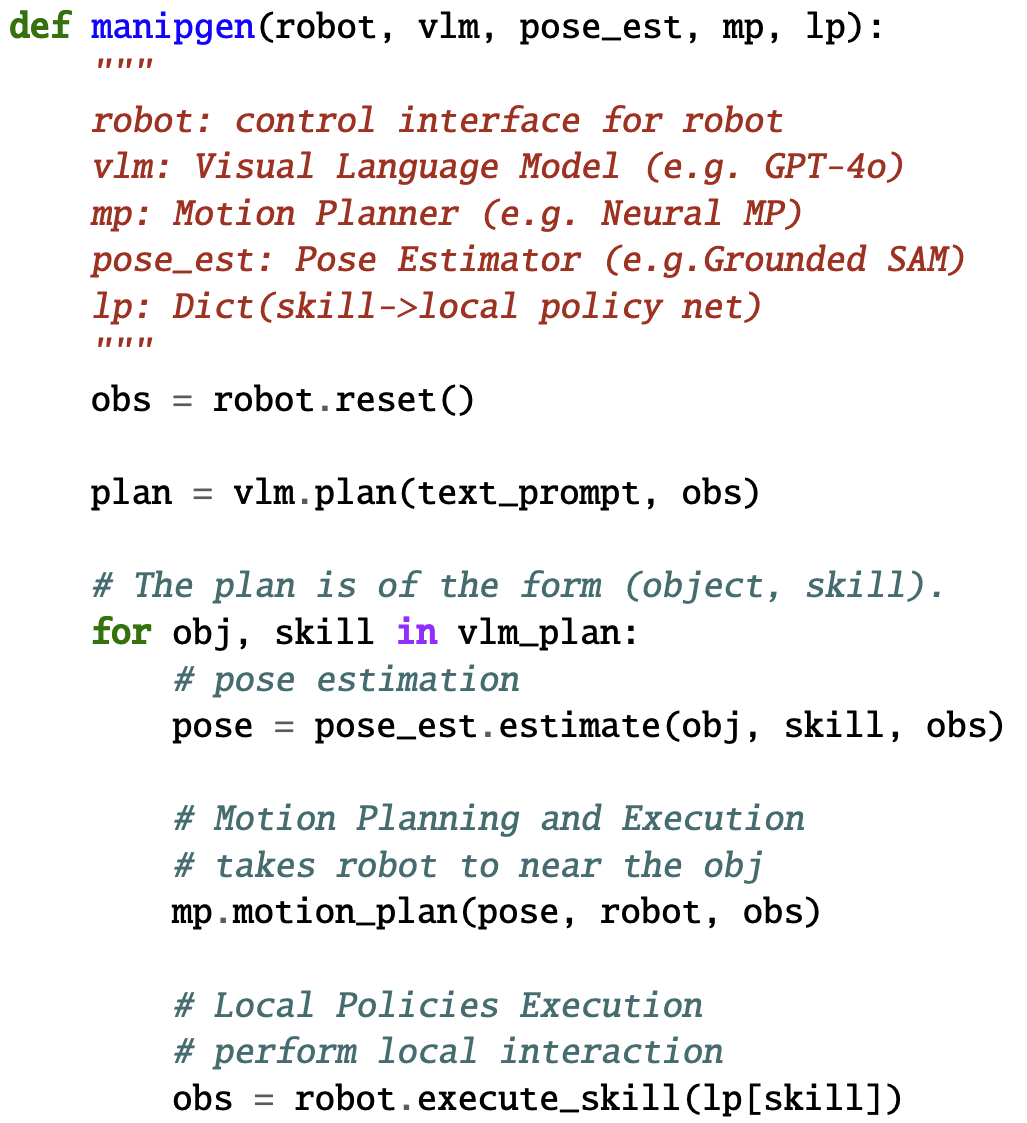}
\end{figure}

\textbf{Tagging} One challenge with Grounded SAM is prompting: the Grounding Dino module is quite sensitive to the input prompt and does not always detect the correct object unless the prompt is formatted well (descriptive, includes colors and texture and shape). Since the VLM will be used to prompt SAM with the target object for each skill, we need the VLM to output tags for objects such that Grounded SAM will trigger on the correct object. As a result, we include an initial tagging phase, in which we have the VLM list all objects that are present in the scene, which we then pass into Grounded SAM to get a list of tags and associated segmentation masks based on how Grounded SAM interprets the scene. We then pass the tagged image, as well as the original image back into the VLM for planning

\textbf{Planning} For the VLM to output a plan (of the same format as in PSL), we provide a system prompt to the VLM that provides it a detailed description of the skill library, their effects on the environment and when they can be used. We also provide it a list of hints as to what consistitutes reasonable plans, this is necessary as existing VLMs still lack strong spatial reasoning capabilities inherently, though prompting seems to alleviate this issue to a certain degree. We include several in-context examples that allow the model to understand the output format, prompt the model to justify its decisions (which helped produce better plans) and format the output as a JSON string (which resolved most of the parsing/formatting issues). 

After planning, ManipGen loops through the plan, estimating the pose of the object/region of interest, motion planning there, and then executing the appropriate local policy. 

\textbf{Pose Estimation} 
We begin by using Grounded-SAM to segment the object of interest, averaging the 3D points of the segmented pixels to determine its position. Orientation is then estimated depending on specific cases. Based on whether there is obstacle above the object, we leverage VLM to classify pick and place tasks into two scenarios: open-space (\textit{e.g.}, table surfaces) and tight-space (\textit{e.g.}, microwaves). We let the robot gripper point downwards in open-space setting. For open-space picking, we project the object’s points onto the XY plane and apply damped least square to fit the points, estimating the object’s longest axis to get gripper's rotation along Z-axis. For open-space placing, we simplify the problem by selecting a fixed orientation pointing down.
In tight-space pick and place, we first sample a set of robot poses around the object. We then capture point cloud of the current scene (excluding the robot) and evaluate the number of points in collision with each sampled pose. To bias towards poses that are further from obstacles, we apply Gaussian noise $\mathcal{N}(0.0, 0.1)$ to the points, and select the pose with minimal collision. For articulated objects, we estimate position of the handle and whether it is vertical or horizontal. Then we sample a set of target poses around the handle and use the same collision-checking method as in tight-space scenario to identify the target pose.

\textbf{Motion Planning}
We use the released Neural MP code and checkpoint to perform motion planning given the target pose and current point-cloud. We perform open-loop motion planning with test-time optimization batch size of 64 (the paper used 100) and max path length of 100. Neural MP can produce paths that are jerky and non-smooth at times. As a result, we perform EMA smoothing with $\alpha=0.9$ to reduce the jerkiness of the trajectories. 

\textbf{Local Policies}
Once initialized near the object of interest, we segment the target object in the first frame using Grounded SAM). We store this mask along with the depth map of the first frame and then we deploy the local policy. At each step, we pass in the segmentation mask of the target object, the first frame depth map, the current frame depth map and the proprioception to the policy. We run the Task Space Impedance Controller on the robot at 60Hz for a fixed, skill specific duration (max 8s). Then, depending on the skill, we either open or close the gripper and begin executing the next stage.

\section{Experiment Details}

\textbf{Hardware} For all of our experiments, we use a Franka Emika Panda Robot, which is a 7 degree of freedom manipulator arm. We control the robot using the Industreallib library (\href{https://github.com/NVlabs/industreallib}{https://github.com/NVlabs/industreallib}) using the Task Space Impedance Controller. For deployment, we use Leaky PLAI with action scales, thresholds and skill deployment durations chosen per skill (Table~\ref{tab:real skill configs}). For all skills, we used a position gain of 1000 and rotation gain of 50.

\begin{table}[H]
    \centering
    \resizebox{\linewidth}{!}{%
    \begin{tabular}{c|c|c|c|c|c}
    \toprule[1.2pt]
        & Pick & Place & Grasp Handle & Open & Close \\\hline
       Duration & 5s & 4s & 6s & 8s & 4.5s \\
       Action Scale Pos & .002 & .002 & .002 & .003 & .005 \\
       Action Scale Rot & .05  & .05 & .004 & .0005 & .0005 \\ 
       Leaky PLAI Thresh X, Y & .02 & .02 & .02 & 0.02 & 0.03 \\
       Leaky PLAI Thresh Z & .02 & .02 & .02 & .003 & .003 \\
       Leaky PLAI Thresh Rot (Deg) & 4 & 4 & 2 & .005 & .005 \\
    \bottomrule[1.2pt]
    \end{tabular}
    }
    \caption{Configurations for skill deployment.}
    \label{tab:real skill configs}
\end{table}

The robot is mounted to a fixed base pedestal behind a desk of size .762m by 1.22m with variable height. For global views (used for the VLM, pose estimation and Neural MP), we use four calibrated depth cameras, Intel Realsense 455, placed around the scene in order to accurately capture the environment. 
We project the depth maps from each camera into 3D and combine the individual point-clouds into a single scene representation for Neural MP and pose estimation.
For input to Neural MP, we further process the point-clouds according to the paper~\cite{dalal2024neuralmp}. While the VLM and the pose estimators can take in multiple views in principle, in practice we found that their results (predicted plans, pose estimations) were significantly less reliable and consistent when using more than a single camera. As a result, for each task, we select one out of the four global view cameras to use for plan prediction and pose estimation. Finally, for local views, we use the d405 camera mounted on the wrist, and pass its depth maps (after clamping and normalization) as input to the policy.

\subsection{Simulated Comparison (Robosuite) Details}
We zero-shot transfer our local policies (trained in IsaacGym) to the robosuite tasks. Note, we do not train on the Robosuite objects at all, we use our data generation and training pipeline to train local policies in IsaacGym and transfer the policies to Robosuite. To deploy our method, we modify the environment to use the UMI~\cite{chi2024universal} gripper and Task Space Impedance Control. We use the same LLM-planning and motion planning infrastructure as used in the PSL~\cite{dalal2024psl} paper and evaluate our policies using 100 trials per task. All other numbers for baselines are taken from the PSL paper.

\subsection{Furniture Bench Experiment Details}
We replicate the exact task setup from the Transic~\cite{jiang2024transic} paper: 3D printed objects from FurnitureBench~\cite{heo2023furniturebench} in the exact poses specified in their paper. In this experiment, we train local policies for these specific tasks and then deploy them. As a result, this experiment compares our method of sim2real transfer (local policies) with end-to-end sim2real transfer (w/ and w/o data aug) as well as Transic (which uses real-world data). 

\subsection{Real World Long-horizon Manipulation Details}

\begin{figure*} [t!]
    \centering
    \subfloat[Cook]{
        \includegraphics[width=0.24\linewidth]{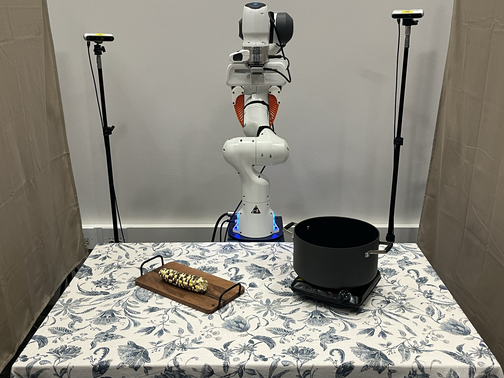}
        \includegraphics[width=0.24\linewidth]{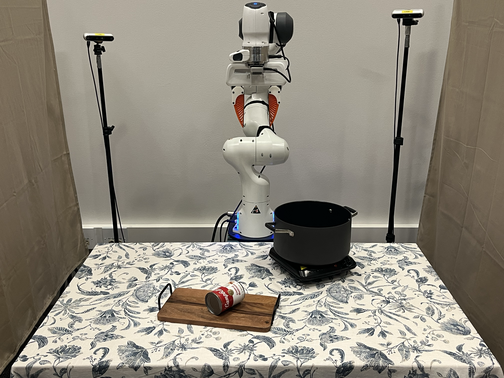}
        \includegraphics[width=0.24\linewidth]{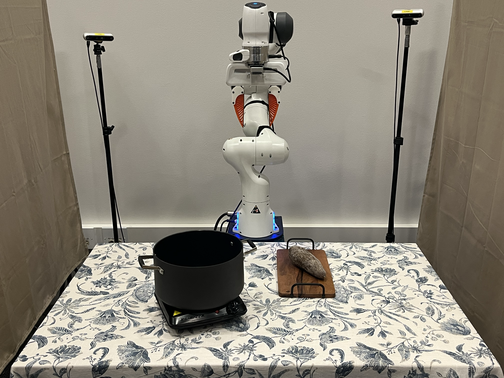}
        \includegraphics[width=0.24\linewidth]{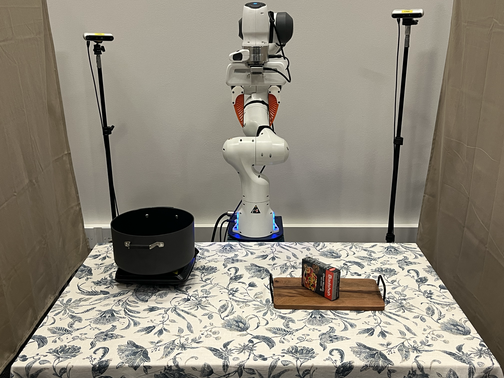}
    }
    \vspace{0.4em}
    \\
    \subfloat[Replace]{
        \includegraphics[width=0.24\linewidth]{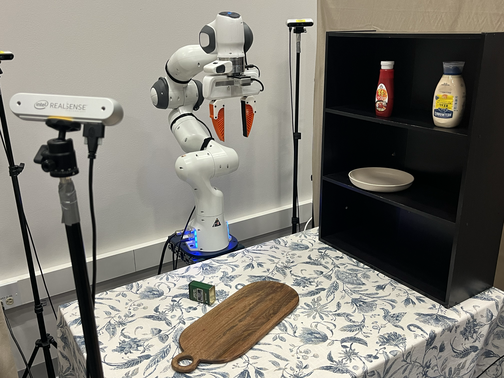}
        \includegraphics[width=0.24\linewidth]{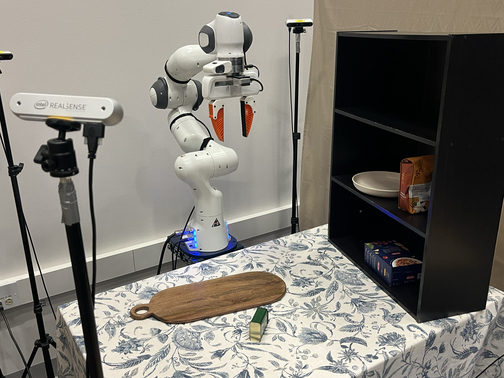}
        \includegraphics[width=0.24\linewidth]{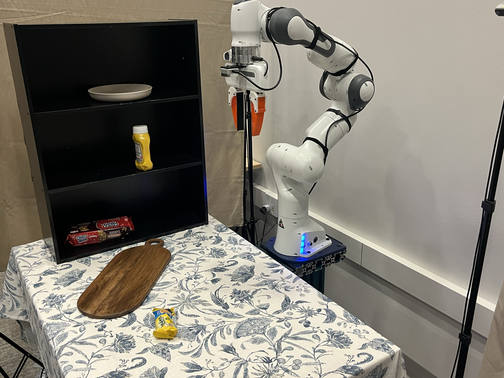}
        \includegraphics[width=0.24\linewidth]{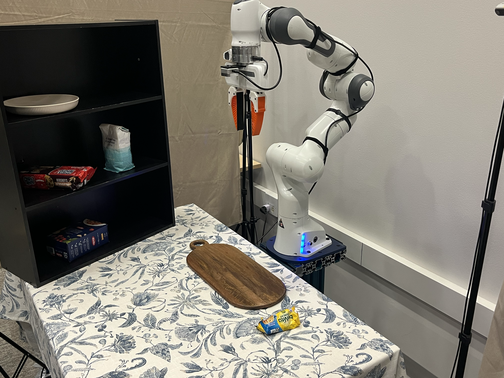}
    }
    \vspace{0.4em}
    \\
    \subfloat[Cabinet Store]{
        \includegraphics[width=0.24\linewidth]{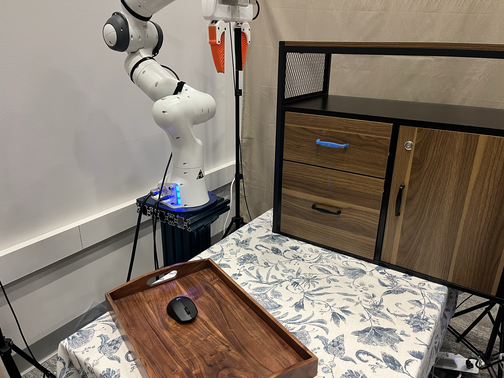}
        \includegraphics[width=0.24\linewidth]{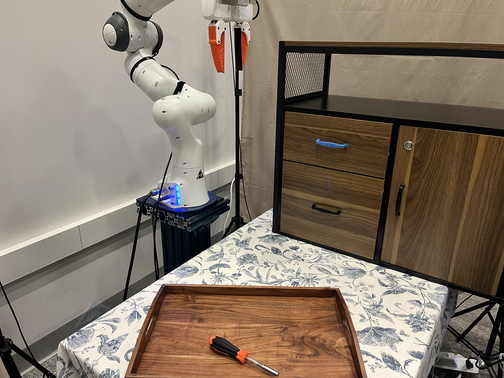}
        \includegraphics[width=0.24\linewidth]{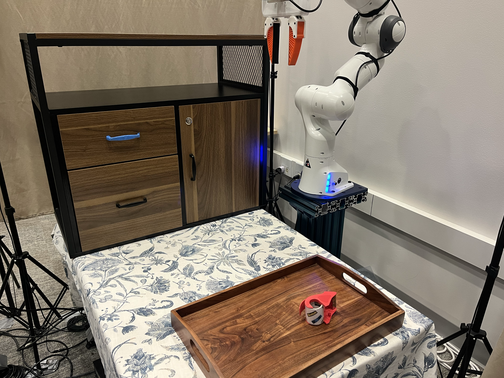}
        \includegraphics[width=0.24\linewidth]{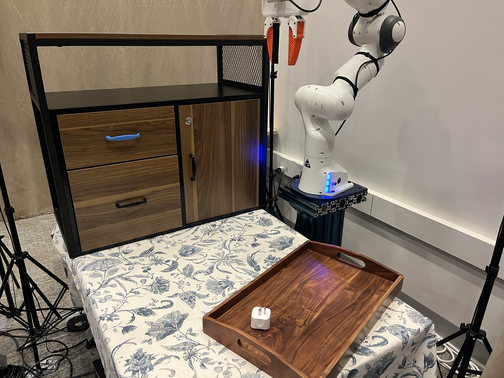}
    }
    \vspace{0.4em}
    \\
    \subfloat[Drawer Store]{
        \includegraphics[width=0.24\linewidth]{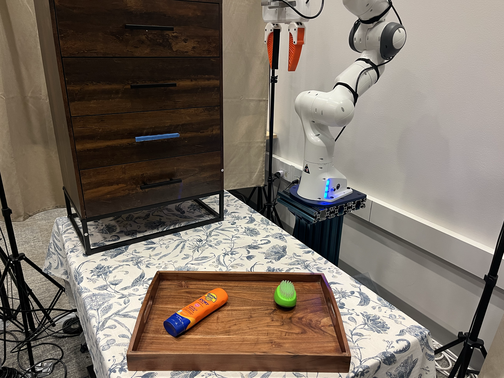}
        \includegraphics[width=0.24\linewidth]{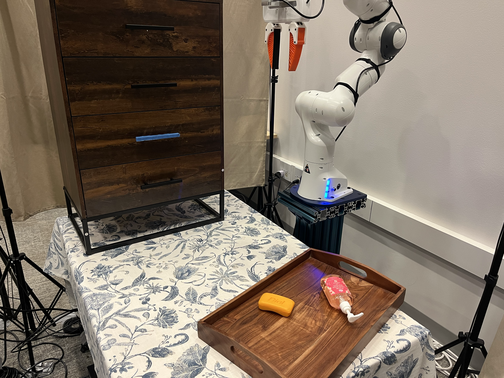}
        \includegraphics[width=0.24\linewidth]{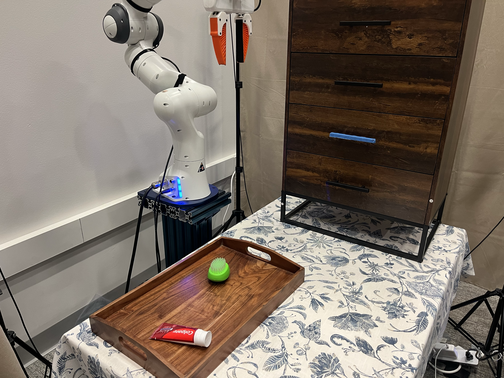}
        \includegraphics[width=0.24\linewidth]{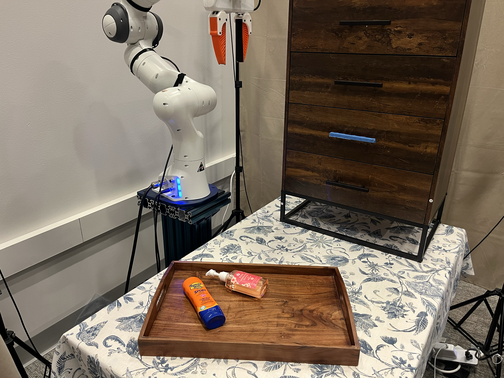}
    }
    \vspace{0.4em}
    \\
    \subfloat[Tidy]{
        \includegraphics[width=0.24\linewidth]{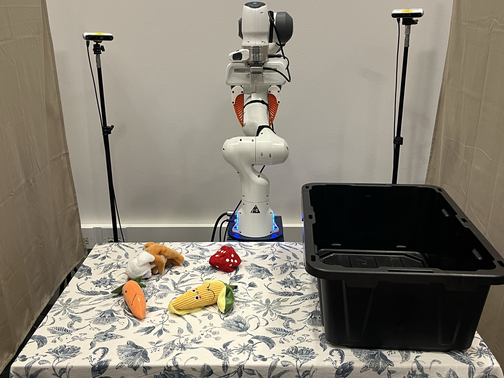}
        \includegraphics[width=0.24\linewidth]{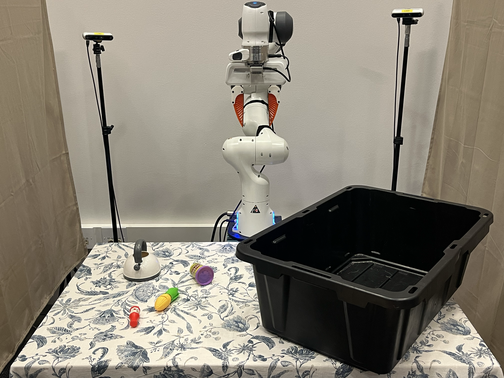}
        \includegraphics[width=0.24\linewidth]{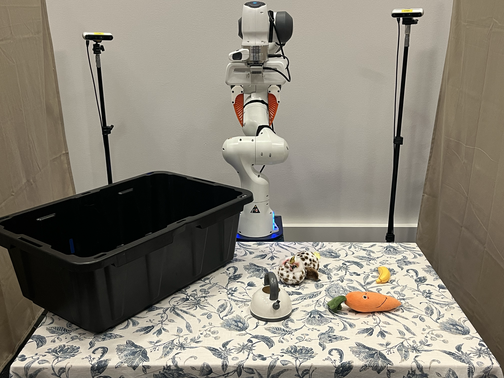}
        \includegraphics[width=0.24\linewidth]{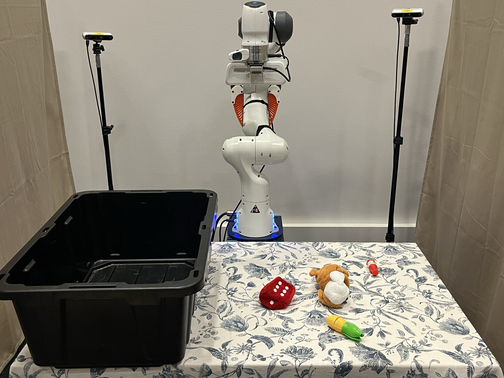}
    }
    \caption{
    Example scene layouts for real world evaluation.
    }
    \label{fig:real_eval_images} 
\end{figure*}

For each of the following tasks, we specify the large object (if present) in each task (as a receptacle) as well as the list of objects we randomize, the task description in detail, and the task prompt provided to ManipGen and the baselines.
Unless otherwise stated, all objects (handles for articulated objects) to interact with are positioned within 0.8 meters of the robot base to ensure they are within the gripper’s reach.
Some example scene arrangements are presented in Fig.~\ref{fig:real_eval_images}.

\subsubsection{Cook (2 stages)} Pick up a food item on the cutting board and put it in a pot on the stove.

Task prompt: \textcolor{ForestGreen}{Put [OBJ] in the black pot.}

Large object: black pot (39cm $\times$ 32cm $\times$ 14cm)

Randomized objects: carrot (17.0cm $\times $ 2.5cm $\times $ 2.5cm), cassava (20.0cm $\times$ 6.1cm $\times$ 6.0cm), corn (17.7cm $\times$ 4.2cm $\times$ 4.3cm), spice box (12.8cm $\times$ 9.0cm $\times$ 3.0cm), soup can (10.0cm  $\times$ 6.7cm $\times$ 6.7cm)

Randomization: We put the food item on a cutting board, the pot on a stove, and randomize their poses across the table within the gripper's reach.

\subsubsection{Replace (4 stages)} Fetch a pantry item from the shelf, put it on a wooden board on the table and take an object from the table, put it on a white plate.

Task prompt: \textcolor{ForestGreen}{Place [OBJ A] on the wooden board, and put [OBJ B] on the white plate in the shelf.}

Large objects: wooden board (51cm $\times$ 18.8cm $\times$ 1.2cm), shelf (80cm $\times$ 60cm $\times$ 23cm)

Randomized objects: [OBJ A] brown coffee package (15.7cm $\times$ 8.0cm $\times$ 6.0cm), blue coffee package (15.7cm $\times$ 8.0cm $\times$ 6.0cm), ketchup bottle (17.5cm $\times$ 9.3cm $\times$ 5.5cm), mustard bottle (17.5cm $\times$ 9.3cm $\times$ 5.5cm), gochujang bottle (17.4cm $\times$ 6.6cm $\times$ 4.1cm); [OBJ B] spice jar (8.0cm $\times$ 4.0cm $\times$ 4.0cm), pepper container (8.2cm $\times$ 5.8cm $\times$ 3.1cm), biscuit pack (10.5cm $\times$ 5.6cm $\times$ 2.5cm)

Randomization: The shelf is placed on the left or right end of the table with randomized orientation between 0 and 30 degrees. [OBJ A] and the white plate are randomly placed on the second or third level of the shelf. The wooden board and [OBJ B] are randomly placed on the other half of the table. We ensure that center of the board and [OBJ B] are at least 30cm away from base of the shelf.

\subsubsection{CabinetStore (4 stages)} Open the drawer with blue handle in the cabinet, put
an office supply inside, and close the drawer.

Task Prompt: \textcolor{ForestGreen}{Store [OBJ] in the drawer with blue handle.}

Large objects: cabinet (80cm $\times$ 71cm $\times$ 40cm, \url{https://www.amazon.com/gp/product/B0CHHTZ52F/ref=ewc_pr_img_12?smid=A38QU35WKLBIKI&psc=1})

Randomized objects: computer mouse (10.5cm $\times$ 6.5cm  $\times$ 3.8cm), tape (10.6cm $\times$ 6.4cm $\times$ 6.3cm), screw driver (18.8cm $\times$ 3.2cm $\times$ 3.2cm), plug (4.9cm $\times$ 4.9cm $\times$ 5.4cm), staple container (10.1cm $\times$ 5.9cm $\times$ 4.0cm)

Randomization: The cabinet is placed on the left or right end of the table, with its back aligned to the table edge. The office supply is randomly placed on a wooden tray on the other half of the table. We ensure that after fully opening the drawer, the office supply is at least 20cm from edge of the drawer.

\subsubsection{DrawerStore (6 stages)} Open a drawer with blue handle, put two personal care items inside, and close the drawer.

Task Prompt: \textcolor{ForestGreen}{Arrange [OBJ A] and [OBJ B] in the drawer with blue handle.}

Large objects: drawer (80cm x 60cm x 23cm, \url{https://www.amazon.com/gp/product/B0BJPLBSHQ/ref=ewc_pr_img_1?smid=A2XKE81PYMCHT4&psc=1})

Randomized objects: brush (8.4cm $\times$ 6.4cm $\times$ 5.4cm), sunscreen bottle (18.9cm $\times$ 5.7cm $\times$  3.8cm), soap (9.4cm $\times$ 5.4cm $\times$ 2.5cm), toothpaste (14.1cm $\times$ 5.7cm $\times$ 3.6cm), sanitizer bottle (17.0cm $\times$ 6.8cm $\times$ 4.5cm)

Randomization: The drawer is placed on the left or right end of the table, with its back aligned to the table edge. We randomly select two personal care items and place them on a wooden tray on the other half of the table. We ensure that after fully opening the drawer, the personal care items are at least 20cm from edge of the drawer.

\subsubsection{Tidy (8 stages)} Clean up the table by putting 4 toys on the table into a bin.

Task Prompt: \textcolor{ForestGreen}{Sort all the toys into the black bin.}

Large objects: black bin

Randomized objects: stuffed carrot (17.2cm $\times$ 4.5cm $\times$ 4.5cm),  stuffed corn (21.3cm $\times$ 5.7cm $\times$ 6.0cm),  stuffed owl (15.0cm $\times$ 7.6cm  $\times$ 6.6cm),  stuffed dog (18.7cm $\times$ 7.8cm $\times$ 8.4cm),  stuffed dice (6.6cm $\times$ 6.6cm  $\times$  6.6cm),  tiny bottle (7.4cm $\times$ 2.9cm $\times$ 2.9cm),  toy teapot (15.1cm  $\times$  12.5cm  $\times$  12.5cm),  toy banana (8.8cm $\times$ 2.9cm $\times$ 4.0cm),  toy corn (12.0cm $\times$ 3.8cm $\times$ 3.8cm), play-doh container (7.6cm $\times$ 6.3cm $\times$ 6.3cm)

Randomization: The bin is placed on the left, right, or front end of the table with randomized orientation between 0 and 180 degrees. The toys are scattered on the other half of the table. We ensure the toys are at least 20cm away from edge of the bin.

\clearpage